\documentclass{article}

\usepackage{amsmath}
\usepackage{amssymb}

\usepackage{graphicx}
\usepackage{url}
\usepackage{algorithm2e}
\usepackage{tikz}
\usetikzlibrary{arrows.meta}

\usepackage{hyperref}
\hypersetup{
  colorlinks   = true,    
  urlcolor     = blue,    
  linkcolor    = red,    
  citecolor    = red      
}
\usepackage{bookmark}
\usepackage{natbib}
\usepackage[capitalize, noabbrev]{cleveref}

\newcommand{\A}{\mathcal{A}}

\makeatletter
\newcommand*{\declarecommand}{%
  \@star@or@long\declare@command
}
\newcommand*{\declare@command}[1]{%
  \provide@command{#1}{}%
  \renew@command{#1}%
}
\makeatother
\newcommand{\myrvspace}[1]{\mathcal{#1}}
\newcommand{\myrvestspace}[1]{\hat{\myrvspace{#1}}}
\newcommand{\myrvest}[1]{\hat{#1}}
\newcommand{\myrvestdist}[1]{q_{\myrvest{#1}}}

\newcommand{\myrvparamrvtheta}[1]{{\Theta_{\myrvest{#1}}}}
\newcommand{\myrvparamrvphi}[1]{{\Phi_{\myrvest{#1}}}}
\newcommand{\myrvparamrvxi}[1]{{\Xi_{\myrvest{#1}}}}
\newcommand{\myrvparamtheta}[1]{{\theta_{\myrvest{#1}}}}
\newcommand{\myrvparamphi}[1]{{\phi_{\myrvest{#1}}}}
\newcommand{\myrvparamxi}[1]{{\xi_{\myrvest{#1}}}}

\newcommand{\myrvparamspacetheta}[1]{{\Delta_{\myrvparamrvtheta{#1}}}}
\newcommand{\myrvparamspacephi}[1]{{\Delta_{\myrvparamrvphi{#1}}}}
\newcommand{\myrvparamspacexi}[1]{{\Delta_{\myrvparamrvxi{#1}}}}

\makeatletter
\newcommand*\defrvar[1]{
  \expandafter\declarecommand\csname s#1\endcsname[1][]{\myrvspace{{#1}}}
  \expandafter\declarecommand\csname h#1\endcsname[1][]{\myrvest{{#1}}}
  \expandafter\declarecommand\csname sh#1\endcsname[1][]{\myrvestspace{#1}}
  \expandafter\declarecommand\csname hp#1\endcsname[1][]{\myrvestdist{#1}}
  \expandafter\declarecommand\csname r#1\endcsname[1][]{\myrvestdist{#1}}
  \expandafter\declarecommand\csname Theta#1\endcsname[1][]{{\myrvparamrvtheta{#1}}}
  \expandafter\declarecommand\csname Phi#1\endcsname[1][]{{\myrvparamrvphi{#1}}}  
  \expandafter\declarecommand\csname Xi#1\endcsname[1][]{{\myrvparamrvxi{#1}}} 
  \expandafter\declarecommand\csname theta#1\endcsname[1][]{{\myrvparamtheta{#1}}}
  \expandafter\declarecommand\csname phi#1\endcsname[1][]{{\myrvparamphi{#1}}}
  \expandafter\declarecommand\csname xi#1\endcsname[1][]{{\myrvparamxi{#1}}}
  \expandafter\declarecommand\csname sTheta#1\endcsname[1][]{\myrvparamspacetheta{#1}}
  \expandafter\declarecommand\csname sPhi#1\endcsname[1][]{\myrvparamspacephi{#1}}
  \expandafter\declarecommand\csname sXi#1\endcsname[1][]{\myrvparamspacexi{#1}}  %
}

\newcommand*\defrvars[1]{
  \@for\@i:=#1\do{\expandafter\defrvar\expandafter{\@i}}}

\newcommand*\defsmallcapsrvar[1]{
  \expandafter\declarecommand\csname h#1\endcsname[1][]{\myrvest{#1}}
}

\newcommand*\defsmallcapsrvars[1]{
  \@for\@i:=#1\do{\expandafter\defsmallcapsrvar\expandafter{\@i}}}  
\makeatother

\defrvars{X,Y,Z,E,S,A,M} %
\defsmallcapsrvars{x,y,z,e,s,a,m}
\newcommand{\pt}{{\prec t}}

\newcommand{\allht}{{0:\hT}}

\newcommand{\thT}{{t:\hT}}

\newcommand{\phis}{\phi^*}

\newcommand{\thetase}{{\theta^1}}
\newcommand{\thetaeae}{{\theta^2}}
\newcommand{\thetae}{{\theta^3}}

\newcommand{\Thetase}{{\Theta^1}}
\newcommand{\Thetaeae}{{\Theta^2}}
\newcommand{\Thetae}{{\Theta^3}}

\newcommand{\piv}{\text{s}}

\newcommand{\hT}{{\hat{T}}}

\newcommand{\mot}{\mathfrak{M}}

\DeclareMathOperator{\p}{{p}}
\DeclareMathOperator{\q}{{q}}

\let\r\relax
\DeclareMathOperator{\r}{{r}}
\let\d\relax
\DeclareMathOperator{\d}{{d}}

\newcommand*{\diff}{\mathop{\kern0pt\mathrm{d}}\!{}}

\DeclareMathOperator*{\argmin}{arg\,min}

\DeclareMathOperator{\KL}{KL}

\DeclareMathOperator{\F}{\mathcal{F}}

\title{Geometry of Friston's active inference}

\author{Martin Biehl\\
          martin@araya.org\\
           Araya Inc.,Tokyo}

\begin{document}

\maketitle

\begin{abstract}
We reconstruct Karl Friston's active inference and give a geometrical interpretation of it. 
%
%
\end{abstract}

\section{Introduction}
\label{sec:intro}

In \citet{biehl_expanding_2018} we have reconstructed the active inference approach as used in \citet{friston_active_2015}. 
Here we present a radically shortened and more general account of active inference. We also present for the first time a geometrical interpretation of active inference.

\label{sec:notation}
We use the same notation and model as in \citet{biehl_expanding_2018}. There readers can also find a translation table to the notations used in \citet{friston_active_2015,friston_active_2016}.

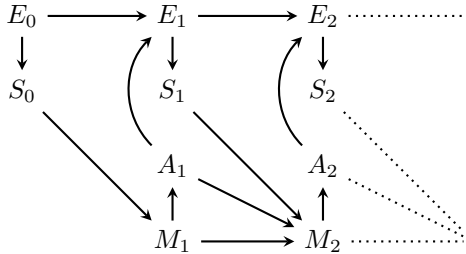
\begin{figure}[ht]
    \label{fig:smloop}%
\begin{center}
    
        \begin{tikzpicture}
    [->,>=stealth,auto,node distance=2cm,
    thick]
    \node (e) [] {$E_1$};
    \node (e') [right of=e] {$E_{2}$};
    \node (s) [below of=e, node distance=1cm] {$S_1$};
    \node (s') [below of=e', node distance=1cm] {$S_{2}$};
    \node (a) [below of=s, node distance=1cm] {$A_1$};
    \node (a') [right of=a] {$A_{2}$};
    \node (m) [below of=a, node distance=1cm] {$M_1$};
    \node (m') [below of=a', node distance=1cm] {$M_{2}$};
    \node (el) [left of=e] {$E_0$};
    \node (er) [right of=e'] {};
    \node (sl) [below of=el, node distance=1cm] {$S_0$};
    \node (sr) [below of=er, node distance=1cm] {};
    \node (al) [below of=sl, node distance=1cm] {};
    \node (ar) [right of=a'] {};
    \node (ml) [below of=al, node distance=1cm] {};
    \node (mr) [below of=ar, node distance=1cm] {};
    \path
      (e) edge node {} (e')
      (e) edge node {} (s)
      (e') edge node {} (s')
      (m) edge node {} (a)
      (m) edge node {} (m')
      (m') edge node {} (a')
      (s) edge node {} (m')
      (a) edge node {} (m')
      (a) edge[bend left=45] node {} (e)   
      (a') edge[bend left=45] node {} (e')
      (el) edge node {} (e)
      (el) edge node {} (sl)
      (sl) edge node {} (m)
      (e') edge[-,dotted] node {} (er)
      (s') edge[-,dotted] node {} (mr)
      (m') edge[-,dotted] node {} (mr)
      (a') edge[-,dotted] node {} (mr) 
      ;
  \end{tikzpicture}
\end{center}
\caption{Bayesian networks of the PA-loop}  
\end{figure}  
\begin{figure}
\label{fig:genmodel}%
\begin{center}
\begin{tikzpicture}      
				[->,>=stealth,auto,node distance=2cm,
		thick]
		\tikzset{
			hv/.style={to path={-| (\tikztotarget)}},
			vh/.style={to path={|- (\tikztotarget)}},
		}
		\tikzset{invi/.style={minimum width=0mm,inner sep=0mm,outer sep=0mm}}		
		\node (e) [] {$\hE_1$};
		\node (e') [right of=e] {$\hE_{2}$};
		\node (s) [below of=e, node distance=1cm] {$\hS_1$};
		\node (s') [below of=e', node distance=1cm] {$\hS_{2}$};
		\node (a) [below of=s, node distance=1cm] {$\hA_1$};
		\node (a') [right of=a] {$\hA_{2}$};
		\node (m') [below of=a', node distance=1cm] {};
		\node (el) [left of=e] {$\hE_0$};
		\node (er) [right of=e', node distance=1cm] {};
		\node (sl) [below of=el, node distance=1cm] {$\hS_0$};
		\node (al) [below of=sl, node distance=1cm] {};
		\node (ar) [right of=a', node distance=1cm] {};
		\node (ml) [below of=al, node distance=1cm] {};
		\node (mr) [below of=ar, node distance=1cm] {};
		\node (th3) [left of=el] {$\Thetae$};
		\node (th2) [above of=th3, node distance=1cm] {$\Thetaeae$};
		\node (th1) [below of=th3, node distance=3cm] {$\Thetase$};
		\node (th2') [above of=e', node distance=1cm] {};  
		\node (th2r) [right of=th2', node distance=1cm] {};   
		\node (c0) [right of=th1, node distance=1cm,invi] {};
		\node (c1) [right of=c0,invi] {};
		\node (c2) [right of=c1,invi] {};
		\node (c3a) [right of=c2,node distance=1cm,invi] {};
		\path
		(th3) edge (el)
		(th2) edge[hv] (e)
		(th2) edge[hv] (e')
		(th2') edge[-,dotted] (th2r)
		(th1) edge[-] (c0)
		(c0) edge[vh] (sl)
		(c0) edge[-] (c1)
		(c1) edge[vh] (s)
		(c1) edge[-] (c2)
		(c2) edge[vh] (s')
		(c2) edge[-] (c3a)
		(c2) edge[-,dotted] (c3a)		
		(e) edge node {} (e')
		(e) edge node {} (s)
		(e') edge node {} (s')
		(a) edge[-,bend left=45,line width=6pt,draw=white] node {} (e)   
		(a') edge[-,bend left=45,line width=6pt,draw=white] node {} (e')
		(a) edge[bend left=45] node {} (e)   
		(a') edge[bend left=45] node {} (e')
		(el) edge node {} (e)
		(el) edge node {} (sl)
		(e') edge[-,dotted] node {} (er)
		(m') edge[-,dotted] node {} (mr)
		;
		\end{tikzpicture}		                 
		\end{center}
		\caption{Bayesian network of the generative model.}
\end{figure}
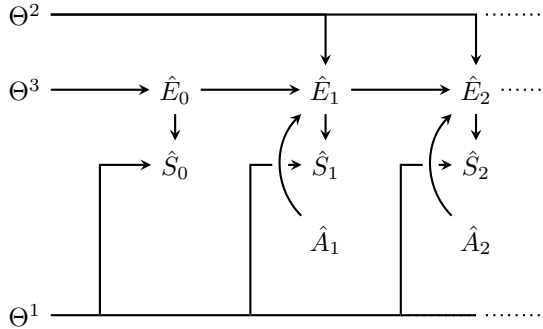

The active inference agent we are describing in the following interacts with an environment according to the perception-action loop defined by the Bayesian network in \cref{fig:smloop}. There, we write $E_t$ for the environment state, $S_t$ for the sensor value, $A_t$ for the action, and $M_t$ for the memory state of the agent at time step $t$. We assume that for all time steps $t$ the transition dynamics of the environment $\p(e_{t+1}|a_{t+1},e_t)$ and the dynamics of the sensors $\p(s_t|e_t)$ are time-homogenous and fixed. 
We further assume that the agent has perfect memory, i.e. at all times $m_t:=sa_\pt$.

The difference to the setup of reinforcement learning for partially observable Markov decision problems is that there is no explicit reward signal. Instead the agent uses a motivation functional $\mot$ which evaluates the agent's current beliefs about the consequences of actions (see \cref{sec:aselectandim}). A standard reward signal $R_t$ can easily be added as an additional sensor value by letting $S_t \rightarrow (S_t,R_t)$.

With the environment, sensor, and memory dynamics fixed it remains to specify the action generation mechanism $\p(a_t|m_t)$. We first describe this in two separate steps, \textit{inference} and \textit{action selection}, that can be performed one after the other. Then we show how active inference combines the two steps in one optimization procedure. On the way we also define motivation functionals.

Action generation in more detail: The agent employs a parameterized model in order to predict the consequences of its actions. At each time step it receives a new sensor value in response to an action and updates its model by conditioning on the new memory state. Additionally conditioning on the various possible future actions (or policies) results in a conditional probability distribution which we call the \textit{active posterior}. The active posterior represents the agent's beliefs about the consequences (for future sensors, latent variables, and internal parameters) of its actions. Obtaining the active posteriors is referred to as the \textit{inference} step. Subsequently, the agent constructs (using softmax) a probability distribution over future actions by assigning high probability to those actions whose entries in the active posterior achieve high values when plugged into the motivation functional (e.g.\ expected future reward if there is an explicit one). This results in the what we call the \textit{induced policy}. 
Obtaining the latter is referred to as the \textit{action selection} step. Afterwards the agent can simply sample from this induced policy to generate actions. 

In active inference, both inference and action selection are turned into optimization problems and then combined in a multiobjective optimization. The inference step can be turned into an optimization using variational inference \citep[see e.g. ][]{bishop_pattern_2011,blei_variational_2017}. Variational inference introduces a variational version of the active posterior. Since the variational active posterior generally differs from the original/true active posterior it leads to a, generally different, \textit{variational induced policy}. Action selection is then formulated as an optimization by introducing an additional (third) policy whose divergence from the variational induced policy is to be minimized. Active inference then optimizes a sum of the respective objective functions and afterwards the agent can sample from the third policy. With a bit of notational trickery this can be written in the form Friston uses, which looks similar to a variational free energy (or evidence lower bound).

\section{Inference}
\label{sec:inference}

\label{sec:genmodel}

The agent performs inference on the generative model given by the Bayesian network in \cref{fig:genmodel}. The variables that model variables occurring outside of $\p(a|m)$ in the perception-action loop (\cref{fig:smloop}), are denoted as hatted versions of their counterparts. 
To clearly distinguish the probabilities defined by the generative model from the true dynamics, we use the symbol $\q$ instead of $\p$. 
Here, $\thetase,\thetaeae,\thetae$ are the parameters of the model.
To save space,write $\theta:=(\thetase,\thetaeae,\thetae)$.

The last modelled time step $\hT$ can be chosen as $\hT=T$ ($T$ is the final step of the PA-loop), but it is also possible to always set it to $\hT=t+n$, in which case $n$ specifies a future time horizon from current time step $t$.

\paragraph{Active posterior}
\label{sec:binference}
\label{sec:plugin}

At time $t$ the agent plugs its experience $sa_\pt$ into its generative model by setting 
$\hA_\tau = a_\tau, \text{for } \tau < t$ 
and 
$\hS_\tau = s_\tau, \text{for } \tau < t$ and conditioning on these.
The consequences of future actions can be obtained by additionally conditioning on each possible future action sequence $\ha_\thT$. This leads to the conditional probability distribution that we call the \textit{active posterior} (the experience $sa_\pt$ is considered fixed):
%
\begin{align}
\label{eq:posteriorgfa}
  \q(\hs_\thT,\he_{0:\hT},\theta|\ha_\thT,sa_\pt)
\end{align}

\paragraph{Variational active posterior}
\label{sec:approxpostandvi}

In active inference the active posterior is obtained via variational inference \citep[see e.g.][]{blei_variational_2017}.

We write $\r$ (instead of $\p$ or $\q$) to indicate variational probability distributions and $\phi$ for the entire set of variational parameters.
Note, the parameter $\phi$ contains parameters for each of the future action sequences $\ha_\thT$ i.e.\
$\phi=(\phi_{\ha_\thT})_{\ha_\thT \in \A^{\hT-t+1}}.$
The variational active posterior is of the form
$\r(\hs_\thT,\he_{0:\hT},\theta|\ha_\thT,\phi)$.
To construct the variational active posterior we cycle through the possible future action sequences $\ha_\thT$ and compute each entry.
For a fixed $\ha_\thT$ the variational free energy, also known as the (negative) evidence lower bound (ELBO) in variational inference literature, is defined as:
\begin{align}
\label{eq:fesimple}
  \begin{split}
    \F[\ha_\thT,&\phi,sa_\pt]:=\\
  &\sum_{\hs_\thT,\he_\allht} \int \r(\hs_\thT,\he_\allht,\theta|\ha_\thT,\phi) \log \frac{\r(\hs_\thT,\he_\allht,\theta|\ha_\thT,\phi)}{\q(s_\pt,\hs_\thT,\he_\allht,\theta|\ha_\thT,a_\pt)} \diff \theta
  \end{split}
\end{align}
Then variational inference amounts to solving for each $\ha_\thT$ the optimisation problem:
\begin{align}
\label{eq:vi}
  \phis_{\ha_\thT,sa_\pt}:=\argmin_\phi \F[\ha_\thT,\phi,sa_\pt].
\end{align}

\section{Action selection and induced policy}
%
\label{sec:aselectandim}

Let $\Delta_{AP}$ be the space of active posteriors. 
Then a \textit{motivation functional} is a map $\mot: \Delta_{AP} \times \A^{\hT-t+1} \rightarrow \mathbb{R}$ taking active posteriors $\d(.,.,.|.) \in \Delta_{AP}$ and a sequences of future actions $\ha_\thT$ to a real value $\mot(\d(.,.,.,|.),\ha_\thT) \in \mathbb{R}$. An example of a motivation functional is the expected value of the sum over future rewards (if one of the sensor values is defined as the reward). Other possibilities can be found in \citet{biehl_expanding_2018}.
Now 
define for some $\gamma \in \mathbb{R}^+_0$ and some motivation functional $\mot$  
a softmax operator $\sigma_\gamma^\mot$ mapping active posteriors $\d(.,.,.|.)$ to probability distributions over future action sequences:
\begin{align}
  \sigma_\gamma^\mot[\d(.,.,.|.)](\ha_\thT):=\frac{1}{Z(\gamma)} e^{\gamma \mot(\d(.,.,.|.),\ha_\thT)}.
\end{align}
Then we call
\begin{align}
  \q(\ha_\thT|sa_\pt):=\sigma_\gamma^\mot[\q(.,.,.|.,sa_\pt)](\ha_\thT)
\end{align}
the \textit{induced policy} of active posterior $\q(.,.,.|.,sa_\pt)$ and $\mot$. And we call
\begin{align}
  \r(\ha_\thT|\phi):=\sigma_\gamma^\mot[\r(.,.,.|.,\phi)](\ha_\thT)
\end{align}
the \textit{induced variational policy} of variational active posterior $\r(.,.,.|.,\phi)$ and $\mot$.

Note that,if $\phi$ is not the optimized parameter $\phis_{sa_\pt}$ then the induced variational policy cannot be expected to lead to actions that actually reflect the preferences encoded in $\mot$. On the other hand, if the true active posterior, or the optimized variational active posterior are used and $\gamma \rightarrow \infty$ the induced policy should correspond to the best guess for an agent with the given generative model, variational distributions, and motivation $\mot$.

\section{Active inference}

\label{sec:activeinference}
%

\begin{figure}
\begin{center}
\begin{tikzpicture}
\coordinate (O) at (0,0,0);

\draw plot [smooth cycle] coordinates {(0.0,0.0)(2.0,1.0)(3,.95)(5,1.2)(6.4,1.0)(6.9,1.0)(6.5,.6)(5.6,.2)(4.8,.0)}; 
\draw        node[circle, fill=black, inner sep=1.5pt, label=below:{$\q(.,.,.|.,sa_\pt)$}] (q) at (1.8,.6) {};
\draw   node[circle, fill=black, inner sep=1.5pt, label=below:{$\r(.,.,.|.,\phi)$}] (r) at (4.8,.8) {};
\draw        node[circle, fill=black, inner sep=1.5pt, label=below:{$\piv(\ha_\thT|\rho)$}] (s) at (7.0,5.0) {};
\draw   node[circle, fill=black, inner sep=1.5pt, label=left:{$\r(\ha_\thT|\phi)$}] (fr) at (4.3,3.8) {};
\draw   node[circle, fill=black, inner sep=1.5pt, label=right:{$\q(\ha_\thT|sa_\pt)$}] (fq) at (2.8,4.8) {};

\draw[-, dashed] (q) edge[out=-8, in=178] node[ fill=white] {$D_1$} (r);
\draw[-, dashed] (fr) edge[out=+28, in=180] node[ fill=white] {$D_2$} (s);
\draw[-{Triangle}] (q) edge[out=+95, in=230] node[fill=white] {$\sigma_\gamma^\mot$} (fq);
\draw[-{Triangle}] (r) edge[out=+115, in=265] node[ fill=white] {$\sigma_\gamma^\mot$} (fr);
\draw[color=white, line width=5] plot [smooth cycle] coordinates {(-0.5,3.5)(2.0,5.0)(4,5.3)(6,5.7)(7.4,5.5)(8.9,4.6)(8.5,4.4)(7.6,3.0)(4.8,3.0)}; 
\draw plot [smooth cycle] coordinates {(-0.5,3.5)(2.0,5.0)(4,5.3)(6,5.7)(7.4,5.5)(8.9,4.6)(8.5,4.4)(7.6,3.0)(4.8,3.0)}; 
\end{tikzpicture} 
\caption{On top the space of probility distributions over future action sequences $\Delta_{\shA^{\hT-t+1}}$ on the bottom the space of active posteriors $\Delta_{AP}$. 
}
\label{fig:manifolds}  
\end{center}
\end{figure}
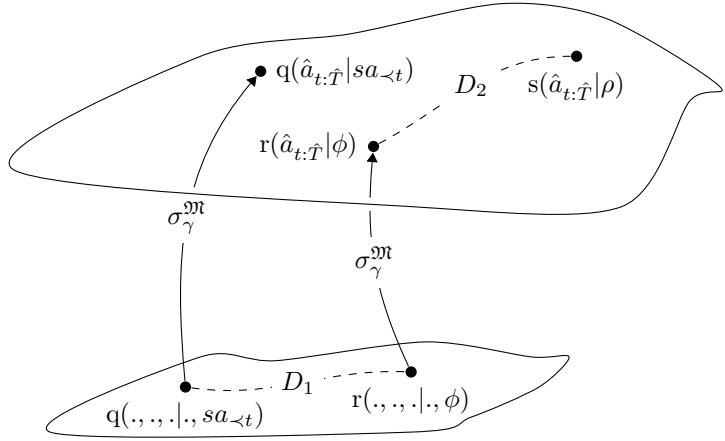

Now introduce an additional variational policy $\piv(\ha_\thT|\rho)$ parameterized by $\rho$ in order to approximate the induced variational policy $\r(\ha_\thT|\phi)$ of the variational posterior (action selection). To make this happen we can minimize the Kullback-Leibler divergence between the two. 

To do inference and action selection at once we can then minimize the sum of the variational free energy (\cref{eq:fesimple}) and the Kullback-Leibler divergence w.r.t. $\rho,\phi$ (for an illustration of the situation see \cref{fig:manifolds}):
\begin{align}
  \underbrace{\sum_{\ha_\thT} \piv(\ha_\thT|\rho) \F[\ha_\thT,\phi,sa_\pt]}_{D_1} + \underbrace{\KL[\piv(\hA_\thT|\rho)||\r(\hA_\thT|\phi)]}_{D_2}
\end{align}
If $\phi$ and $\rho$ are optimized the agent can then sample actions from $\piv(\ha_\thT|\rho)$. 

Now if we change notation and let $\piv(\ha_\thT|\rho) \rightarrow \r(\ha_\thT|\rho)$ and $\r(\ha_\thT|\phi)\rightarrow \q(\ha_\thT|\phi)$ then the above can be rewritten as:
\begin{align}
  \sum_{\ha_\thT,\hs_\thT,\he_\allht} \int \r(\hs_\thT,\he_\allht,\theta,\ha_\thT|\rho,\phi) \log \frac{\r(\hs_\thT,\he_\allht,\theta,\ha_\thT|\rho,\phi)}{\q(s_\pt,\hs_\thT,\he_\allht,\theta,\ha_\thT|\phi,a_\pt)}\diff \theta.
\end{align}
This looks similar to a variational free energy or evidence lower bound and is the form of the free energy found in \citet{friston_active_2016}. What distinguishes it from a true variational free energy is the occurrence of the parameter $\phi$ in numerator and denominator.

\subsubsection*{Acknowledgement}
I want to thank Manuel Baltieri and Thomas Parr for helpful discussions as well as Christian Guckelsberger, Daniel Polani, Christoph Salge, and Sim\'on C. Smith, for feedback on this article.
\bibliographystyle{apalike}
\bibliography{bibliography}

%
%
%
%

\end{document}